\documentclass[conference]{IEEEtran}

\usepackage{url}
\usepackage{amsmath}
\usepackage{amssymb}
\usepackage{graphicx}
\usepackage{mathrsfs} 
\usepackage{booktabs}
\usepackage{cite}
\usepackage{tabularx} % For width of tables
\usepackage{todonotes}
\usepackage{outlines}
\usepackage{xfrac} % imports '\sfrac'

\usepackage{diagbox}
\usepackage{layouts}

\usepackage{float}
\usepackage{subfig}
\usepackage{import}
\usepackage{pgf}
\usepackage{tikz}
\usetikzlibrary{shapes}
\usepackage[utf8]{inputenc}\DeclareUnicodeCharacter{2212}{-}

\makeatletter
\DeclareRobustCommand{\rvdots}{%
  \vbox{
    \baselineskip4\p@\lineskiplimit\z@
    \kern-\p@
    \hbox{.}\hbox{.}\hbox{.}
  }}
\makeatother

\title{Co-Training an Observer and an Evading Target}
\author{Andr\'{e} Brandenburger, Folker Hoffmann, Alexander Charlish \\
	Fraunhofer FKIE\\
	\url{{andre.brandenburger,folker.hoffmann,alexander.charlish}@fkie.fraunhofer.de} \\
}

\date{\today}

\begin{document}
%T-width \printinunitsof{in}\prntlen{\textwidth}\\
%L-width \printinunitsof{in}\prntlen{\linewidth}\\
%T-Height \printinunitsof{in}\prntlen{\textheight}\\

\maketitle

\begin{abstract}
Reinforcement learning (RL) is already widely applied to applications such as 
robotics, but it is only sparsely used in sensor management. In this paper, we 
apply the popular Proximal Policy Optimization (PPO) approach to a multi-agent UAV 
tracking scenario. While recorded data of real scenarios can accurately reflect 
the real world, the required amount of data is not always available. Simulation 
data, however, is typically cheap to generate, but the utilized target behavior is
often naive and only vaguely represents the real world. 
In this paper, we utilize multi-agent RL to 
jointly generate protagonistic and antagonistic policies and overcome the data 
generation problem, as the policies are generated on-the-fly and adapt 
continuously. This way, we are able to clearly outperform baseline methods 
and robustly generate competitive policies. In addition, we investigate 
explainable artificial intelligence (XAI) by interpreting feature saliency and 
generating an easy-to-read decision tree as a simplified policy. %
\end{abstract}

\section{Introduction}

\begin{figure}
	\centering
	\def\svgwidth{\linewidth}
	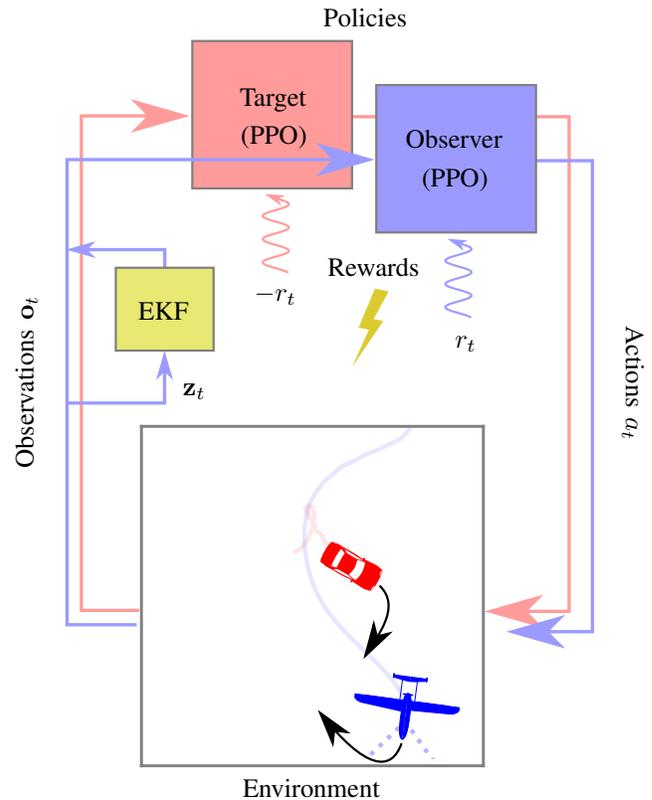
	\caption{A schematic overview of our method. We solve a classical UAV 
control task through reinforcement learning. Two separate policies (top) choose 
actions (right) based on realistic environment observations (left). An EKF is 
utilized to filter noisy sensor measurements. Most importantly, we employ 
contradicting reward signals for the individual policies to innervate 
antagonistic behaviors.}
	\label{fig:schematic}
\end{figure}

Reinforcement learning (RL)~\cite{Sutton2018} offers the promise of learning the behavior of an agent, requiring only the specification of its reward function. 
RL could therefore lead to a generic way to perform sensor management, in which only the sensing objective needs to be defined by the system designer. 
The reinforcement learning algorithm then automatically learns a behavior, called the policy, to fulfill this objective.
In many applications, the agents could theoretically learn their policies online in the real world, however the poor performance during early stages of training and in novel, unseen situations make such an approach unpractical. 
Instead, learning the behavior in a simulated environment and afterwards transferring it to the real world is more feasible. 
To achieve this, the environment must specify the movement of the non-cooperative targets to be tracked. One could use pre-defined, fixed trajectories, which induces the risk of overfitting to those scenarios. Alternatively, it is possible to let the targets move randomly, which is less realistic. In this work, we instead follow the approach of training against a worst-case target, which counteracts the sensor management.
This is achieved by training a target policy with inverted rewards parallel to the sensor management.
The policies are trained using multi-agent reinforcement learning based on observations of the platform states and target estimates.
A schematic overview of the method can be seen in Fig.~\ref{fig:schematic}.

We consider a sensor path planning problem, where the trajectory of an unmanned aerial vehicle (UAV) is optimized. 
The UAV tracks a mobile ground-based target using a range-bearing sensor, which is restricted in its field of view (FOV). 
Such measurements are typical for a radar.
Utilizing path planning to optimize the performance of a sensor is a classical 
problem of sensor management and several algorithms have been proposed, mostly 
based on online trajectory optimization. A common way to evaluate and 
demonstrate these methods is based on pre-defined target trajectories. For example, 
\cite{Hernandez2004} optimizes the trajectories of bearing-only sensors and 
\cite{Zhou2011} optimizes the trajectories of heterogeneous sensor platforms 
containing range and/or bearing sensors. In both cases, results are shown for linearly 
moving targets. 

Evaluating a sensor path planning algorithm on a pre-defined scenario is 
reasonable if the algorithm is guaranteed to also work for other situations. 
However, when using a reinforcement learning algorithm, this imposes the risk 
that the agent behavior overfits to this single 
scenario. Consequently, the policy would only perform well on the given 
scenario. Therefore, existing works in path planning using reinforcement learning, have modeled the target behavior as random. The
work in \cite{Engin2020} optimizes the path of a range-only or bearing-only 
sensor and models the targets as either stationary or following a 2-D Brownian 
motion model. The work in \cite{Bhagat2020} considers a target in an urban 
context, which moves randomly on a partially occluded road grid. The observer 
learns a behavior to always keep the target in its field of view. In 
\cite{Hoffmann2020}, the authors train a policy to localize stationary targets, 
which are placed randomly based on a prior. We note that some of the works based 
on online optimization also evaluate their policies using random target 
behavior~\cite{Miller2009}. Commonly, this random behavior does not follow a specific intent of the target, but instead is based on a fixed probability distribution on the action space.

However, the assumption that a target moves randomly without intent is 
often not met in practice. The targets typically have some intent in their 
behavior. An alternative to the random model would be to find real data based on 
targets behavior, on which a policy is trained. This would require a large 
amount of data to avoid overfitting on specific trajectories. In addition, 
targets might behave differently when the observer policy changes. Lastly, 
real-world data is mostly difficult and costly to acquire.

Alternatively, a game-theoretic approach can be taken. Instead of modeling 
specific target trajectories, we assume that the target has the intent to maximally degrade the tracking performance. 
When training an observer policy under this worst-case assumption, 
we can expect to achieve better tracking performance for other target behaviors. 
Such a worst-case target is known as an evading target and has 
been considered at several places in the literature for path planning based on 
online optimization. The work in \cite{Ragi2013} includes, among others, an 
evasive target model. The target knows the position of the tracking UAVs and 
always moves away from the closest observer. A more elaborate avoidance model is 
used in \cite{Theodorakopoulos2009}, where the ground-based target optimizes its 
trajectory to hide from an observer between obstacles. The observer takes this 
target behavior into account when optimizing its own trajectory.

Still, optimizing with respect to a hard-coded evasive target model can lead to overfitting.
In this work, we therefore take the approach of training the policy of the evading 
target in parallel to the policy of the tracking UAV. The target is 
another agent, whose goal is to deteriorate the tracking performance of the observer 
as far as possible. If the observer policy specializes too much on the current 
policy of the target, the target could ideally learn that a change in its 
behavior leads to the observer tracking it less accurately and, consequently, choose 
another behavior. 

Extending the field of RL to multiple agents is called multi-agent reinforcement 
learning. In this setting, the policy learned by each agent not only depends on the 
environment, but also on the learned policies of the other agents. As the other agents 
might have different goals, the policy of each agent needs to take the policies 
of the other agents into account. When these other agents improve in a competitive setting, this 
leads to a successively increasing difficulty for an agent to achieve its own 
goals. This feedback can be interpreted as a form of curriculum 
learning~\cite{Narvekar2020}. Co-training of an agent and its antagonist has led 
to several noteworthy breakthroughs in recent years, especially in the form of 
self-play, where the policy plays against a potentially different version of 
itself. Exemplary applications are to learn Go~\cite{Silver2016} or 
Starcraft~\cite{Vinyals2019}. 

In this paper, we formulate a setting related to pursuit-evasion problems. In these tasks, 
a single pursuer or a group needs to catch one or multiple evading targets by 
reaching their position. Several solutions to this problem are based on explicitly modeling 
the agent-behavior~\cite{Janosov2017,Li2015a,Angelani2012}. Recently, learning 
based solutions have been investigated~\cite{Souza2020, Lowe2017}. Training an 
additional policy for the evading targets can lead to a complex co-evolution of 
strategies~\cite{Baker2020}. 
While addressing a similar application, the problem studied in this paper 
varies from the pursuit-evasion category mentioned in previous work.
In the traditional pursuit-evasion setting, pursuers are required to reach the position of the target. In contrary, 
this paper addresses a sensor management problem, where a pursuer needs to achieve optimal 
measurement geometry towards the evading target. The goal is not to reach the target 
position, but instead to optimally localize it.

In this paper, we do not consider the actual transfer from the simulation to a real system. 
This adds additional complexities next to the behavior of the target, as the sensor model 
and the movement model of all platforms must correspond to the real system. Such a sim-to-real 
transfer is actively researched in the reinforcement learning and robotics communities, using 
techniques like domain randomization with promising results \cite{Peng2018,openai2019solving}.

In this paper, we apply multi-agent reinforcement learning to the problem of 
tracking an evading target. In Section~\ref{sec:method} we describe the 
tracking approach, the sensor management problem and the training method. In 
Section~\ref{sec:results} we show simulative results and explanations for the 
trained policies. Finally, Section~\ref{sec:conclusion} concludes the paper.

\section{Method}
\label{sec:method}
\subsection{Multi-Agent Reinforcement Learning}
Due to its direct relationship to the trained agent behavior, the simulation environment 
has to be carefully designed and parameterized. In reinforcement learning, the 
environment is commonly modelled as a Markov decision process (MDP) and 
can be described as a tuple $\left(S,A,P,R,\gamma\right)$. More precisely, it 
consists of the environment state $S$, action space $A$, state transition 
probabilities $P: S \times A \times S \to \left[0,1\right]$, reward function $R: 
S\times A \to \mathbb{R}$ and discount factor $\gamma\in \left[0,1\right]$. 
Typically, agents cannot observe the environment state directly, but only 
through a projection $\Omega : S \to O$ for an observation space $O$. 
Altogether, this yields a Partially Observable Markov Decision Process (POMDP), 
defined as $\left(S,A,P,R,\gamma, O, \Omega \right)$.

This environment can be seen as an interaction framework for an agent that seeks 
to maximize accumulated rewards (or \textit{return}) by choosing propitious actions $a_t \in A$ from an action space $A$. 
More specifically, this manifests in a mapping $\Pi: O \to A$, called the 
\textit{policy}. The policy is trained while accounting for all discounted future rewards, enabling the agent to learn long-term behaviors. 
Traditionally, only a single agent is taken into account. In 
our application, we consider a multi-agent environment, where two agents, the 
observer $\mathcal{O}$ and the target $\mathcal{T}$ can interact with each other. 
Consequently, we will define two different sets of observations $O_\mathcal{O}, 
O_\mathcal{T}$, actions $A_\mathcal{O}, A_\mathcal{T}$ and rewards 
$R_\mathcal{O}, R_\mathcal{T}$ for the respective agents.

For more detailed information on modelling RL environments, refer to \cite{Sutton2018}. 

\subsection{Tracking Framework}
In this paper, we address the task of trajectory planning for two unmanned 
platforms. We model one unmanned aerial vehicle (UAV) and one ground-based 
vehicle. For the sake of simplicity, we consider the task only in two dimensions and 
neglect the altitude of the platforms. The corresponding states of the agents 
$\mathcal{A}\in\left\{\mathcal{O}, \mathcal{T}\right\}$ are parameterized by 
their position $\mathbf{x}_{\mathcal{A},t}\in\mathbb{R}^2$, rotation 
$\phi_{\mathcal{A},t}\in\left[-\pi, \pi\right]$, forward speed
$v_{\mathcal{A},t}\in\mathbb{R}$ and rotational speed 
$\omega_{\mathcal{A},t}\in\mathbb{R}$. To reduce the complexity of the 
optimization, we assume $v_{\mathcal{A},t}$ to be constant, whereas 
$\omega_{\mathcal{A},t}$ can be used to steer the vehicle.
The simulation frequency is defined as $\tau > 0$, meaning at each step 
$\sfrac{1}{\tau}$ seconds elapse.  At every time-step $t 
\mapsto t+1$, the platform state is updated according to classical mechanics
\begin{align}
 \mathbf{x}_{\mathcal{A},t+1} &= \mathbf{x}_{\mathcal{A},t} + \frac{1}{\tau}
\left(\begin{array}{r}
\text{cos}\left(\phi_{\mathcal{A},t}\right) \\ 
-\text{sin}\left(\phi_{\mathcal{A},t}\right) \\
\end{array}\right) v_{\mathcal{A},t} \\
\phi_{\mathcal{A},t+1} &= \phi_{\mathcal{A},t} + \frac{1}{\tau}\omega_{\mathcal{A},t}\,.
\end{align}

To model the observer $\mathcal{O}$ that senses the target platform 
$\mathcal{T}$, we simulate a radar sensor on the observer platform. This radar 
yields noisy local measurements $\mathbf{z}_{\mathcal{T},t}$ of the target relative to 
the observer in polar coordinates, such that
\begin{align}
    \Delta \mathbf{x}_t &= \mathbf{x}_{\mathcal{O},t} - \mathbf{x}_{\mathcal{T},t}\\
	d_{t} &= \vert\vert \Delta \mathbf{x}_t \vert\vert_2\\
	\theta_{\mathcal{T},t} &= \text{atan2}\left( (\Delta \mathbf{x}_t)_y, (\Delta \mathbf{x}_t)_x \right) - \phi_{\mathcal{O},t}\\
	\mathbf{z}_{\mathcal{T},t} &\sim 
	\mathcal{N}\left(
		\left[
			d_{t},
			\theta_{\mathcal{T},t}
		\right]
		;
		\text{diag}
		\left(
			\sigma_d^2,
			\sigma_\theta^2
		\right)
	\right)\,.
\end{align}
The front-facing sensor is restricted to a fixed FOV with constant opening $0 < \alpha 
\leq 2 \pi$, where a measurement is only registered iff 
$\vert\theta_{\mathcal{T},t}\vert < \frac{1}{2}\alpha$. Additionally, we define 
the number of time steps since the last registered measurement as $\Delta t_{t,z}$.

We employ an extended Kalman filter (EKF) to reconstruct a target 
estimate from the noisy sensor measurements. The EKF utilizes the true sensor 
noise matrix and the popular piecewise constant white acceleration dynamics 
model~\cite{bar2004_dwpa}. This results in the target track $\left(\left[\mathbf{\tilde{x}}_{\mathcal{T}, t}, 
\mathbf{\tilde{v}}_{\mathcal{T}, t}\right], \mathbf{P}_{\mathcal{T}, t}\right)$, which 
is an estimate of the target position and velocity with corresponding covariance 
$\mathbf{P}_{\mathcal{T}, t}$ in the global frame. 

Similar to $d_{ t}$ and $\theta_{\mathcal{T}, 
t}$, we define 
\begin{align}
    \Delta \mathbf{\tilde{x}}_t &= \mathbf{x}_{\mathcal{O},t} - \mathbf{\tilde{x}}_{\mathcal{T},t}\\
	\tilde{d}_{t} &= \vert\vert 
	\Delta \mathbf{\tilde{x}}_t \vert\vert_2\\
	\tilde{\theta}_{\mathcal{T},t} &= \text{atan2}\left( (\Delta \mathbf{\tilde{x}}_t)_y, (\Delta \mathbf{\tilde{x}}_t)_x \right) - \phi_{\mathcal{O},t}\,,
\end{align}
based on the constructed track.

\subsection{Agent Models} 
\begin{figure}
 \begin{center}
		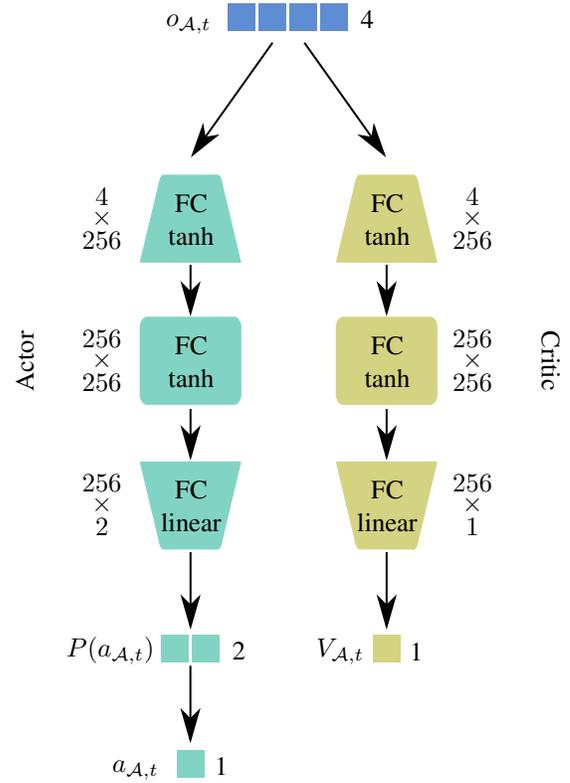
	\end{center} 
	\caption{The model architecture for the learned policies.
On the left, the actor network is shown, which learns 
the parameterization of a gaussian action distribution. On the
right we display the critic network. The models utilize fully 
connected (FC) layers with \textit{tanh} activation for the hidden 
units and linear activation for the output units.}
    \label{fig:ppomodel}
\end{figure}

For our method, we utilize the Proximal Policy Optimization 
(PPO) algorithm~\cite{Schulman2017}, which has proven as highly performant in a wide 
range of tasks such as UAV control~\cite{Koch2020, Hoffmann2020}. A schematic of 
the network architecture is shown in Fig.~\ref{fig:ppomodel}. As seen before, 
the use of RL methods requires the careful definition of the observation- and 
action space, as well as a reward function that reflects the goal of the optimization.

\subsubsection{Observation Spaces}
To enable us to supply the agents with a scalar indicator of uncertainty instead 
of the multi-dimensional covariance, we calculate the entropy
\begin{equation}
	S_t = \frac{1}{2}\ln \det \left( 2 \pi e 
	\mathbf{P}_{\mathcal{T}, t} \right) \,.
\end{equation}

In addition, we process the time since the last measurement $\Delta t_{t,z}$ 
using a decay function
\begin{equation}
 \kappa_t = k^{-\Delta t_{t,z}}\,, 
\end{equation}
for a $k > 1$. In contrast to $\Delta t_{t,z}$, $\kappa_t$ is bounded, which is 
beneficial for training stability.

\begin{figure}
	\centering
	\def\svgwidth{\linewidth}
	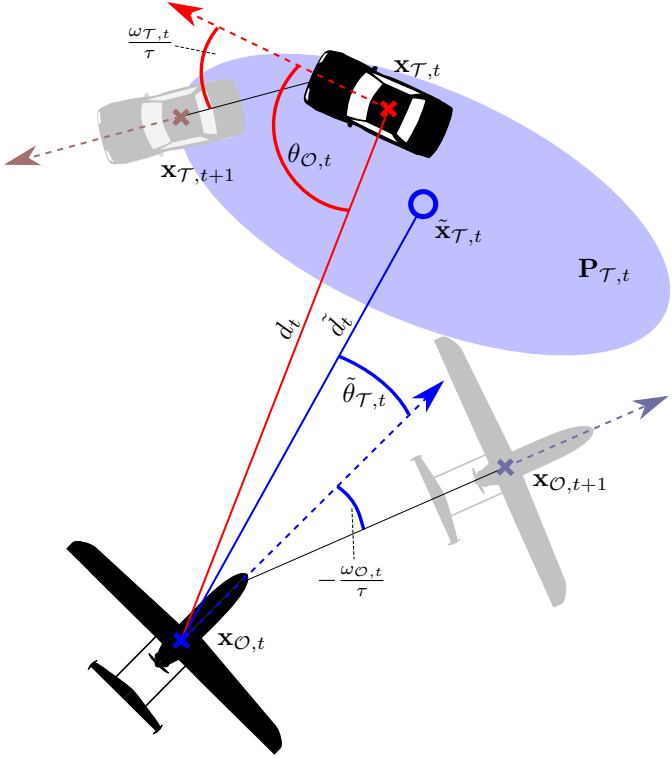
	\caption{A partial visualization of the observation space. The UAV (bottom left) builds an estimate $\tilde{\mathbf{x}}_{\mathcal{T}, t}$ of the
		 target (top center) with corresponding covariance $\mathbf{P}_{\mathcal{T}, t}$. This estimate is transformed into local polar 
		 coordinates $( \tilde{d}_{\mathcal{T}, t}, \tilde{\theta}_{\mathcal{T}, t} )$ before it is added to the observation vector. 
		Similarly, the target observation uses a polar coordinate representation of the true UAV position.
		Both platforms change heading with respect to their rotational velocities $\omega_{\mathcal{O}, t}$ and $\omega_{\mathcal{T}, t}$. }
	\label{fig:obs_schematic}
\end{figure}

Accordingly, we summarize the environment state by projecting to the agent observation
\begin{align}
	\mathbf{o}_{\mathcal{O}, t} &= \left[ 
	\omega_{\mathcal{O}, t}, 
	\kappa_t, 
	\tilde{d}_{t}, 
	\text{sin}\left(\tilde{\theta}_{\mathcal{T},t}\right),
	\text{cos}\left(\tilde{\theta}_{\mathcal{T},t}\right),
	S_t
	\right] \\
	\mathbf{o}_{\mathcal{T}, t} &= \left[
	\omega_{\mathcal{T}, t},
	d_{t}, 
	\text{sin}\left(\theta_{\mathcal{O},t}\right),
	\text{cos}\left(\theta_{\mathcal{O},t}\right)
	\right]\,,
\end{align}
with $\theta_{\mathcal{O},t}$ defined analogously to
$\theta_{\mathcal{T},t}$. Both angles are decomposed into sine and cosine terms, correcting the discontinuity of the angle representation at $\pm \pi$.
A partial illustration of the observation space is shown in Fig.~\ref{fig:obs_schematic}.

This way, the observations yield a concise 
representation of the environment state. For the observer, this not 
only includes information about the targets position and platform velocity, but 
also the time-since-measurement decay $\kappa_t$ and the uncertainty measure 
$S_t$. While $S_t$ is crucial for assessing the track uncertainty, $\kappa_t$ 
provides a notion of time to the otherwise time-agnostic observer, allowing for 
more complex behaviors. While the observer is based on the estimated target position, the target is given the true position of the observer.
Note that, the goal in this paper is to train an observer policy, which is robust against maneuvering and evading targets.
By making the worst case assumption of the target knowing the true position of the observer, the target is strengthened in its role as the antagonist. 

Note that, by normalizing the observations $o_{\mathcal{A}, t}$ with the limits 
$d_{\max}$ and $S_{\max}$, it is possible to achieve $o_{\mathcal{A}, t} \in 
[-1, 1]$. Since normalized values are typically easier to learn by neural 
networks, we utilize this augmentation to increase the training performance.
The parameter $d_{\max}$ denotes the maximal range of interest, and $S_{\max}$ a value empirically higher than any encountered track entropy.

\subsubsection{Action Spaces}
To enable the agents to navigate in the environment, we accept actions 
$a_{\mathcal{A},t} \in \left[ -1, 1 \right]$ that steer the platform, while 
complying with velocity limits $\omega_{\mathcal{A},\text{min}}, 
\omega_{\mathcal{A},\text{max}}$, such that
\begin{gather}
 \omega_{\mathcal{A},t} = 
    \frac{1}{2} \left( \omega_{\mathcal{A},\text{max}} - \omega_{\mathcal{A},\text{min}} \right) 
    \left(a_{\mathcal{A},t} + 1\right)
    + \omega_{\mathcal{A},\text{min}}\,.
\end{gather}
Thus, both agents are able to adapt their rotational velocities to control their 
movement in the environment.

\subsubsection{Reward Functions}
The reward function captures the optimization goals of the RL agents. Since we 
consider a sensor management task, the observer will maximize the tracking 
performance, which we formulate as minimization of the track entropy $S_t$. On 
the contrary, the target will attempt to counteract the observer as well as 
possible. 
Consequently, we define the rewards as
\begin{align}
	r_{\mathcal{O},t} &= 1-\frac{S_t}{S_{\max}}\\
	r_{\mathcal{T},t} &= - r_{\mathcal{O},t} \,.
\end{align}
This way, we set conflicting objectives for the observer $\mathcal{O}$ and the 
target $\mathcal{T}$, since the observer will try to increase the track 
accuracy, whereas the target aims to reduce it. 

\subsubsection{Termination Criteria}
\label{sec:termination}
A single simulation run is ended if any of the following termination criteria
\begin{align}
 t &> t_{\max} & d_{t} &> d_{\max} & \tilde{d}_{t} &> d_{\max} \,,
 \label{eq:termination}
\end{align}
is fulfilled, with a maximum run length $t_{\max}$ and maximum target distance $d_{\max}$. 
As common in the reinforcement learning literature, we denote such a single simulation run as an episode.
These termination criteria contribute to richer training data, as uninteresting 
situations are not recorded. This mainly comes into effect during early stages 
of training, where the policies have not yet captured their respective goals. 
Recognize, that negative rewards give an incentive to terminate, whereas 
positive rewards encourage the continuation of episodes. Therefore, we ensure 
that $r_{\mathcal{O},t} > 0$ and $r_{\mathcal{T},t} < 0$ hold. This causes the 
observer to prefer long episodes, whereas the target is inclined to end the 
episode, if possible.

\section{Experimental Results}
\label{sec:results}
Firstly, we define simple baseline methods, which are used as a 
comparison for the performance of our approach. As a naive agent, a random policy 
is constructed, where actions $a_t \in \left[-1,1\right]$ are drawn uniformly at random:
\begin{align}
	a_{\mathcal{O}_R,t} &\sim U\left(-1,1\right) \\
	a_{\mathcal{T}_R,t} &\sim U\left(-1,1\right)
\end{align}
Secondly, we define proportional control policies (P-Controller), which based on the angular 
deviations $\theta_{\mathcal{A},t}$ and gains $K_\mathcal{A}$, steer the 
observer towards the target and the target away from the observer:
\begin{align}
	a_{\mathcal{O}_P,t} &= K_\mathcal{O} 
	\frac{\theta_{\mathcal{T},t}}{\pi} \\
	a_{\mathcal{T}_P,t} &= \text{sgn}\left( \theta_{\mathcal{O},t} \right) K_\mathcal{T} 
	\left(1 - \frac{\vert \theta_{\mathcal{O},t} \vert}{\pi} \right)
\end{align}
This way, we achieve simple baseline behaviors for evasion and tracking. 
Note that, those policies are based on the ground truth direction towards the target. 
This makes the baseline stronger and assures that it does not fail in situations, where the track is inaccurate.

Furthermore, in the course of this evaluation, we use the following 
environment parameterization:
\begin{align*}
 \alpha &= 1.4\,\text{rad} & \sigma_d &= 40\,\text{m} & \sigma_\theta &= 0.005\,\text{rad}\\
 k &= 1.05 & S_\text{max} &= 50 & d_\text{max} &= 5000\,\text{m}\\
 t_\text{max} &= 20\,\text{min} & \tau &= 0.5\,\text{Hz} & \gamma &= 0.99  \\
 \omega_{\mathcal{O},\text{min}} &= -0.25\,\text{rad} & \omega_{\mathcal{O},\text{max}} &= 0.25\,\text{rad} & v_{\mathcal{O}, \text{x}}&= 50\,\sfrac{\text{m}}{\text{s}}\\
 \omega_{\mathcal{T},\text{min}} &= -0.75\,\text{rad} & \omega_{\mathcal{T},\text{max}} &= 0.75\,\text{rad} & v_{\mathcal{T}, \text{x}}&= 20\,\sfrac{\text{m}}{\text{s}}\\
 K_\mathcal{O} &= 1.0 & K_\mathcal{T} &= 0.2 &&
\end{align*}

\subsection{Performance}
We train the model on a standard work station machine, with a \textit{AMD Ryzen 7 PRO 3700U}, utilizing 3 CPUs and no GPU. 
It is expected that inference should also be possible on a smaller, embedded system.
In total, the training over $0.8\text{M}$ environment 
interactions takes approximately $45$ minutes. Our implementation is based on 
\textit{Ray RLLib}\cite{liang2018rllib} with a TensorFlow backend, which allows 
for high scalability if needed, reducing the training time even further. 

\begin{figure}
	\begin{center}
		\input{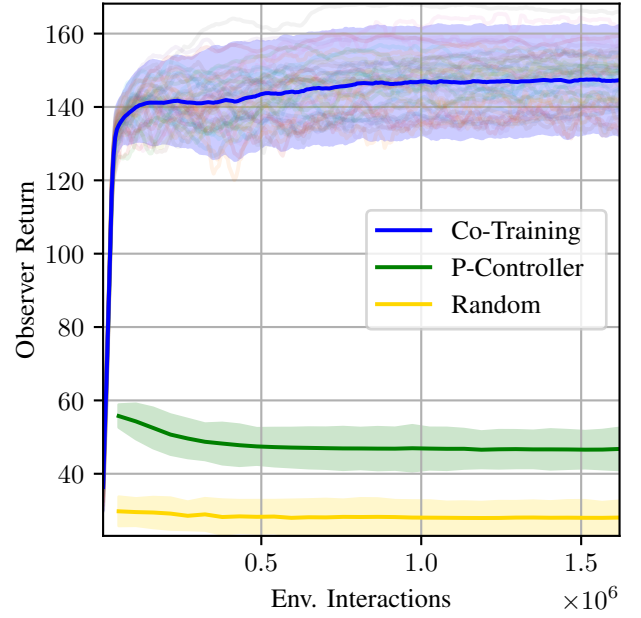}
	\end{center}
	\caption{Observer learning curves. The returns of the individual training 
runs for the proposed approach are shown as the shaded lines and the mean 
performance as the opaque blue line. The opaque green line shows the mean 
performance of the P-Controller observer, whereas the random policy is 
displayed in yellow. The shaded regions correspond to $\mu \pm2\sigma$.
	}
	\label{fig:training_curve}
\end{figure}

To validate network convergence, we have performed $50$ training runs with 
approximately $1.5\text{M}$ environment steps. The resulting observer returns 
during training are shown in Fig.~\ref{fig:training_curve}. Note that, the mean 
observer performance only marginally increases after $0.8\text{M}$ steps, 
indicating that the policy has converged. As seen in 
Fig~\ref{fig:training_curve}, our model not only outperforms the random agent, 
but also consistently exceeds the performance of the P-Controller for all 
trained policies. It can be seen that, since the target policy is trained in 
conjunction with the observer, the difficulty of the task increases over time. 
This is clearly reflected in the P-Controller performance, which decreases over 
the course of the training. 

In Tables~\ref{table:baselines} and \ref{table:baselines_std} we compare the 
performance and stability of the baseline methods to our trained agents at $0.8\text{M}$ 
steps. As already shown in Fig.~\ref{fig:training_curve}, our approach clearly 
outperforms the simpler baseline solutions. Observe in Table~\ref{table:baselines}, however, that the best 
agent against a random observer is the P-Controller target. As the best strategy 
against an incapable observer is to simply maximize the distance to the observer 
and end the episode, the P-Controller performs better in that scenario. This 
shows that our approach implicitly assumes a capable opponent policy. 
Nevertheless, this is a valid assumption for most real world applications. 
Furthermore, we show low mean performance variance for fixed policies in 
Table~\ref{table:baselines_std}.

\begin{table}[!t]
	\centering
	\begin{tabular}{c|c|c|c}
		\diagbox{Target}{Observer} & Co-Training & P-controller & Random \\
		\hline
		Co-Training & $146.34 \pm 15.1$ & $46.90 \pm 5.6$ & $28.21 \pm 5.1$ \\
		P-controller & $153.79 \pm 39.6$ & $130.93 \pm 0.0$ & $22.29 \pm 0.0$ \\
		Random & $151.74 \pm 21.6$ & $54.44 \pm 0.0$ & $28.51 \pm 0.0$ \\
	\end{tabular}
	\caption{Mean observer episode return $\mu \pm 2\sigma$ compared to baseline 
		methods after $0.8\text{M}$ training steps. The standard deviation $\pm 2\sigma$ 
		indicates the mean performance variation after re-training the model from 
		scratch. Note that, the first row corresponds to the results of 
		Fig.~\ref{fig:training_curve}.
		Since the standard deviation is calculated over the different 
		training runs, $\sigma=0$ holds when comparing the static baseline methods
	}
	\label{table:baselines}
	
	\vspace{1em}
	\centering
	\begin{tabular}{c|c|c|c}
		\diagbox{Target}{Observer} & Co-Training & P-controller & Random \\
		\hline
		Co-Training & $\pm 10.7$ & $\pm 10.4$ & $\pm 37.0$ \\
		P-controller & $\pm 11.9$ & $\pm 5.6$ & $\pm 29.8$ \\
		Random & $\pm 6.6$ & $\pm 18.7$ & $\pm 37.9$ \\
	\end{tabular}
	\caption{Mean observer episode return standard deviations $\pm 2\sigma$ 
		after $0.8\text{M}$ training steps. The standard deviation $\pm 2\sigma$ shows 
		the mean performance deviation of fixed policies over different test episodes. 
		Consequently, the standard deviation indicates how much the return of a single trained 
		policy fluctuates during different inference episodes.
	}
	\label{table:baselines_std}
\end{table}

In addition, we assess the mean optimal subpattern assignment (OSPA) metric \cite{Schuhmacher2008} over the different models. 
This metric combines information on the track accuracy with the number of correctly tracked targets.
We compute the average OSPA metric over episodes with a fixed duration $t_{\text{max}}$. 
According to Eq.~\ref{eq:termination}, a single episode terminates at time $t_{\text{term}} < t_{\max}$ if the target can be considered lost by the observer.
Thus, if an episode terminates early, we view the target track as dropped for the remaining episode.
This allows us to reformulate the OSPA metric for a single target as
\begin{equation}
		\frac{1}{t_{\text{max}} \tau}
		\left(
		\sum_{t}^{t_{\text{term}}}
		\left[
			\min(\vert\vert \mathbf{x}_{\mathcal{T}, t} - \tilde{\mathbf{x}}_{\mathcal{T}, t}\vert\vert_2, c)
		\right]
		+ (t_{\text{max}}-t_{\text{term}}) \tau c
		\right)\,,
\end{equation}
with a cutoff distance $c$. In contrast to the return, the OSPA metric is a direct indicator of the 
tracking performance. Analogous to previous evaluations, the OSPA metric is computed over $50$ 
training runs after $0.8\text{M}$ steps of training, each of which is evaluated over $50$ test episodes.
In Table~\ref{table:ospa} it can be seen that the Co-Training method also outperforms
 the baseline approaches with respect to this more general metric.

\begin{table}[!t]
	\centering
	\begin{tabular}{c|c|c|c}
	
	\diagbox{Target}{Observer} & Co-Training & P-controller & Random \\
	\hline
	Co-Training & $271 \pm 94$ & $425 \pm 28$ & $471 \pm 25$ \\
	P-controller & $158 \pm 93$ & $348 \pm 0$ & $464 \pm 0$ \\
	Random & $194 \pm 50$ & $457 \pm 0$ & $453 \pm 0$ \\
	
	\end{tabular}
	\caption{Comparison of the mean OSPA metric $\mu \pm 2\sigma$ after $0.8\text{M}$ training steps 
		with cutoff distance $c=500\text{m}$. The lower the OSPA metric, the better the localization of the target. 
		Since the standard deviation is calculated over the different 
		training runs, $\sigma=0$ holds when comparing the static baseline methods.}
	\label{table:ospa}
\end{table}

For further investigation, we use the model with median return at 
$0.8\text{M}$ steps. This way, we ensure that representative 
results are achieved without bias towards particularly performant models.

\subsection{Policy Explanation (XAI)}
As more complex artificial intelligence (AI) approaches find their ways into our 
lives, model explanation strongly gains importance. For instance, the reasoning 
of neural network models is opaque by nature, which undermines trust for 
non-experts and makes them hard to employ for security critical systems. There 
are already numerous suggestions on how to explain AI methods, an overview about 
many methods can be found in \cite{barredo2020}. In this paper, we explain 
the learned policy on the basis of two approaches: The ad-hoc analysis of the feature 
importance using saliency and the post-hoc policy approximation using decision 
trees. Note that, although we only evaluate the observer policy, the same 
procedures can be applied to investigate the target policy.

\begin{figure*}
	\begin{center}
		\import{images}{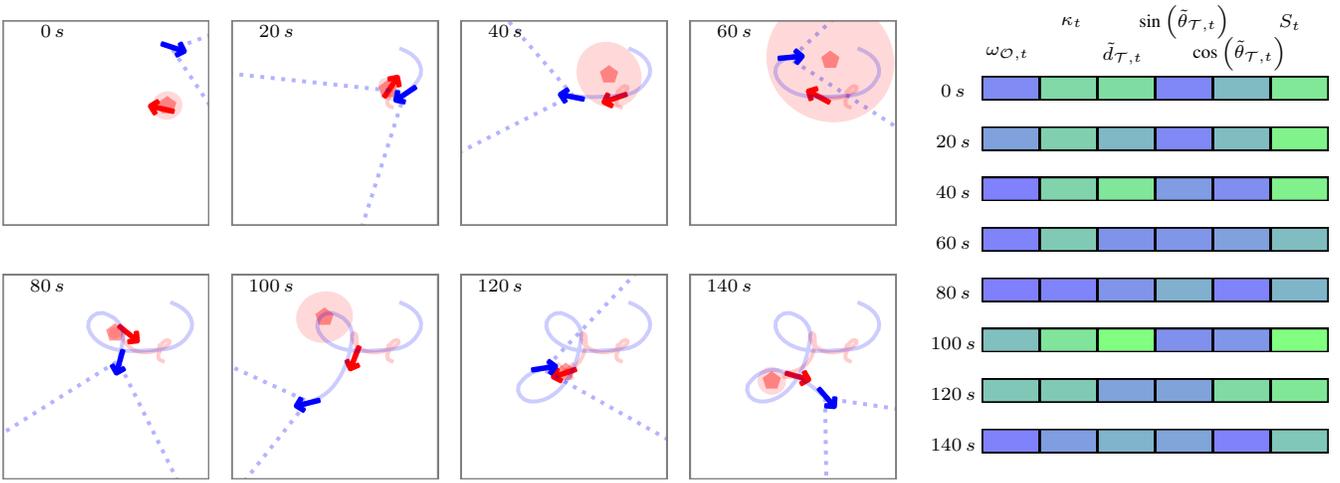}
	\end{center}
	\caption{Behavior of the median policies trained for $0.8\text{M}$ environment interactions
(left) and gradients with respect to the agent observations (right). The blue 
arrow denotes the observer pose, with the field of view as the dashed blue line 
and the full observer trajectory in shaded blue. Analogously, the target is 
shown in red, with the target estimate as the red pentagon with corresponding 
standard deviation $2\sigma$ shown as the red ellipses. On the right, the 
normalized gradients $\frac{da_{\mathcal{O},t}}{d\mathbf{o}_{\mathcal{O},t}}$ are plotted 
for each of the time steps, where blue indicates low and green represents high 
gradients.}
\label{fig:qualitative}
\end{figure*}

Similar to the neural network training mechanism, it is possible to utilize the 
gradients to visualize importance of specific input features given a network 
output. This is particularly popular in computer-vision approaches, where the 
gradients can be processed to \textit{saliency maps}, which can be used to find 
important regions of the input image. In our approach, it is possible to utilize 
the same principle: By plotting $\frac{da_{\mathcal{O},t}}{d\mathbf{o}_{\mathcal{O},t}}$, 
we can assess the sensitivity of the action $a_{\mathcal{O},t}$ with respect to 
the observation $\mathbf{o}_{\mathcal{O},t}$. A sequence of (non-consecutive) frames with 
the corresponding gradients is shown in Fig.~\ref{fig:qualitative}. Note that, 
the observer succeeds in tracking the target, while regularly maneuvering over it and 
taking strong turns to keep it in the observers limited FOV. On the 
other hand, the target apparently follows the observer. Although this might 
appear to be a counter-intuitive strategy to avoid tracking, this policy allows 
the target to exploit the limited FOV of the observer. By keeping the distance 
to the observer small, the target is able to repeatedly find its way behind the 
observer, becoming invisible for the observer and forcing it to take expensive, 
strong turns to relocate the target.

In the plotted normalized gradients, it is apparent, that the most important 
features are $\kappa_t$, $\tilde{d}_{t}$ and $S_{t}$, whereas 
$\omega_{\mathcal{O}, t}$ 
$\text{sin}(\tilde{\theta}_{\mathcal{T},t})$ and 
$\text{cos}(\tilde{\theta}_{\mathcal{T},t})$ are less significant for this 
episode. Note that, this only explains the policy locally. The observations in 
the recorded episodes are not uniformly distributed, but depend on the policy. 
Therefore, an analysis based on those observations is biased towards regions of 
the state space that are frequently visited by the policy. While this local 
explanation can give insight in the policy behavior in the mean case, it is 
often imperative to also assess the behavior in less likely edge cases, which 
will be part of the next technique. 

\begin{figure*}
	\centering
	\begin{tikzpicture}[level distance=8mm]
		\import{images}{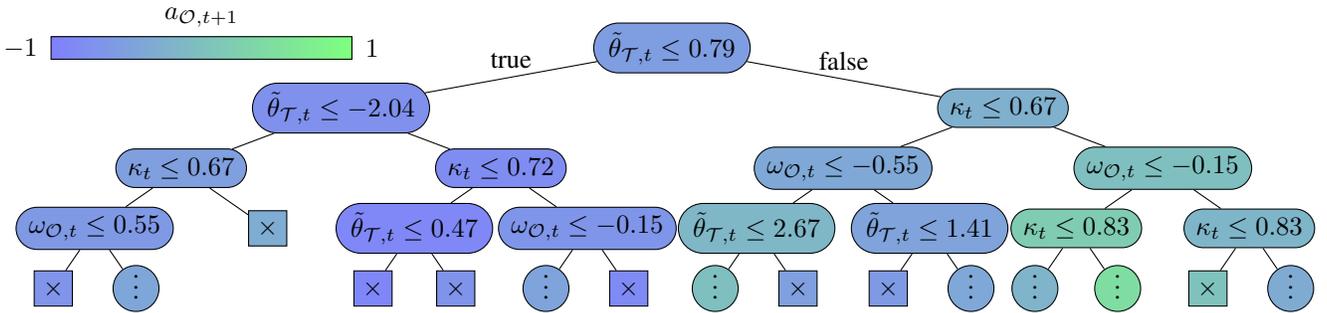}
	\end{tikzpicture}
	\caption{Decision tree of the median-performance model generated using 
artificial observations. The nodes resemble a numerical decision criterion, 
where the color indicates the action that would be taken at that node. By 
traversing the tree downwards, the chosen actions mimic the trained policy 
closer. Vertical dots show truncated splitting nodes, whereas leaf nodes are 
displayed as "$\times$".}
	\label{fig:decision_tree}%
\end{figure*}

To enable global understanding of the policy, we construct a post-hoc decision 
tree model of the observer. By evaluating the agent reactions 
$a_{\mathcal{O},t}$ on a uniform $6$-D grid of observations $\mathbf{o}_{\mathcal{O},t}$, 
it is possible to generate representative samples of the policy $\Pi: O \to A$. In each dimension 
of the observation, we generate $21$ samples. Subsequently, we utilize these 
samples to train a decision tree regressor, while employing pruning to ensure 
generalization and to limit the size of the tree. An excerpt of the resulting tree 
can be seen in Fig.~\ref{fig:decision_tree}, where the color indicates the 
action that would be taken if the tree is evaluated until the corresponding 
node. Note that, the observations 
$\text{sin}(\tilde{\theta}_{\mathcal{T},t})$, 
$\text{cos}(\tilde{\theta}_{\mathcal{T},t})$ are considered jointly for the tree 
generation, as their values are correlated through $\tilde{\theta}$. Generally, negative actions are 
located on the left side of the tree, whereas positive actions tend to be 
produced by the right most nodes. As common with decision trees, nodes in the 
upper levels of the tree can be seen as the most important nodes, whereas lower 
levels only handle details of the task. As expected, the decision tree only 
partially agrees with the saliency analysis. For uniformly sampled observations, 
the track angle $\tilde{\theta}_{\mathcal{T}, t}$ and the last action 
$\omega_{\mathcal{O},t}$ significantly gain importance, whereas the previously 
important track distance $\tilde{d}_{t}$ does not occur at the top 
layers. 

In addition, it is possible to analyze individual paths in the tree to understand the general decision process of the policy. For example, the shown tree produces the strongest turns for the following criteria:
\begin{align*}
 a_{\mathcal{O},t} &\approx -1: & 
 -120^{\circ} < &\tilde{\theta}_{\mathcal{T},t} \leq 27^{\circ} \land \kappa_t \leq 0.72 \\
 a_{\mathcal{O},t} &\approx 1: & 45^{\circ} < &\tilde{\theta}_{\mathcal{T},t} \land 0.83 < \kappa_t \land \omega_{\mathcal{O},t} \leq -0.15
\end{align*}

In the first case, the agent takes a strong right turn if the target estimate is 
to its right, but the target has not been seen for at least $8$ steps (encoded 
in $\kappa_t$). This stands in contrast to the second case, 
where the platform rotated away from the target in the previous step. Now the target is on the left of the platform, slightly outside the observer FOV. 
As the target was seen at most 3 steps ($\kappa_t$) before, it can be expected that a strong left turn moves the target back into the field of view.
Due to the different values of $\kappa_t$, the two 
cases are fundamentally different, since the right turn is induced by the 
absence of information, whereas the left turn is taken due to recent 
measurements. 
However, it must be noted that the decision tree is not an exact approximation of the true policy, due to the coarse per-dimension sample size, as well as inherently due to a finite tree.
Consequently, it can not be expected 
to capture the full potential of the agent, prohibiting the exhaustive 
explanation through this method. Still, the decision tree gives important 
insight into the general decision process of the otherwise opaque neural network 
model.

\section{Conclusion}
\label{sec:conclusion}
In this paper, we have presented an approach to jointly train observer and 
target policies in an UAV tracking scenario. 

The successful implementation of our method opens the door to numerous areas of 
research and applications. For instance, the agents can be utilized to generate 
trajectories, which in turn can be used as training or evaluation data for other 
approaches. Traditionally, generating these type of behavioral trajectories 
often necessitates recording real world data, which can be expensive due to 
factors such as hardware wear and supervision by human experts. While 
hand-crafted simulated trajectories can be cheaper, they are still 
time-consuming to design. With suitable 
reward functions, these costly methods of data acquisition can now be substituted 
by trained agents, which can mirror the competitive behavior of typical 
real-world scenarios. Although finding matching reward functions can be 
challenging, they are mostly significantly easier to formulate than the 
policies themselves. 

Additionally, and most apparently, our models can be directly used for 
controlling UAVs in a tracking scenario with an evading target. Previous 
methods, that were either based on simple behavioral procedures or trained on 
recorded target trajectories, are prone to overfitting and do not properly 
reflect real-world cases. By co-training both agents hand-in-hand, we generate 
the data on-line and thus mitigate these shortcomings. 

Moreover, the method promises to perform well on more challenging tasks, which 
may manifest in higher measurement noise, more mobile targets or occlusion. 
Investigation and adaption on these scenarios poses an interesting foundation 
for future work. Furthermore, it is possible to incorporate a specific sensor model for the target.
This can lead to different behavior, as the observer may learn to exploit 
potential blind spot of the target, deteriorating its evasion capabilities

Additional future work could also analyze the policies in more depth, utilizing 
more sophisticated XAI strategies and exchanging with domain experts. 
Finally, insights of these explanations can be used to refine the model.

\bibliographystyle{IEEEtran}
\bibliography{refs}

\end{document}